\documentclass{Interspeech}
\usepackage{hyperref}
\usepackage{url}
\usepackage{graphicx}   %

\usepackage{subcaption}
\usepackage{comment}
\usepackage{svg}

\interspeechcameraready

\title{ Brain-tuned Speech Models Better Reflect Speech Processing Stages in the Brain}

\author[affiliation={}]{Omer}{Moussa}
\author[affiliation={}]{Mariya}{Toneva}

\affiliation[nocounter]{}{Max Planck Institute for Software Systems}{Germany}
\email{\{omoussa, mtoneva\}@mpi-sws.org}

\usepackage{comment}

\begin{document}

\maketitle

\begin{abstract}
Pretrained self-supervised speech models excel in speech tasks but do not reflect the hierarchy of human speech processing, as they encode rich semantics in middle layers and poor semantics in late layers. Recent work showed that brain-tuning (fine-tuning models using human brain recordings) improves speech models' semantic understanding. Here, we examine how well brain-tuned models further reflect the brain’s intermediate stages of speech processing. We find that late layers of brain-tuned models substantially improve over pretrained models in their alignment with semantic language regions. Further layer-wise probing reveals that early layers remain dedicated to low-level acoustic features, while late layers become the best at complex high-level tasks. These findings show that brain-tuned models not only perform better but also exhibit a well-defined hierarchical processing going from acoustic to semantic representations, making them better model organisms for human speech processing.
\end{abstract}

\section{Introduction}

With the rapid development of self-supervised (SSL) language and speech models, research at the intersection of neuroscience, language, and speech understanding has advanced considerably. Current pretrained self-supervised (SSL) speech and language models can predict human brain responses to natural speech with impressive accuracy, outperforming hand-engineered features \cite{millet2022toward, vaidya2022self, antonello2024scaling, oota2023speech}. Researchers seek to use language models as model organisms for studying reading and listening in the brain \cite{toneva2021bridging}, aiming to better understand the information processing mechanisms that give rise to brain-like representations.

Prior work has also shown that the ability of these models to predict brain responses (i.e., their ``brain alignment'') differs substantially across their layers \cite{wehbe2014aligning, jain2018incorporating,toneva2019interpreting,schrimpf2021neural,caucheteux2020language, vaidya2022self, antonello2024scaling}. For SSL speech models specifically,  upper-middle layers align much more with the whole cortex response and the auditory cortex than other model layers \cite{vaidya2022self}. Some semantic hierarchy is also observed in downstream tasks, where low-level and spectral features are best captured by earlier layers, while phonemic and semantic tasks are best captured by upper-middle layers \cite{vaidya2022self}. 

However, these analyses of how semantics evolve through model layers did not test how these models align with the brain's high-level late language regions, which are thought to be the endpoint of linguistic and speech processing in the brain \cite{deniz2019representation}. Moreover, SSL models show a downward trend after the upper-middle layers on high-level tasks, suggesting that the late layers (the ones closest to the model output) encode semantics less effectively than the preceding ones. We propose that a model that is more in line with speech processing in the brain would exhibit a better-defined hierarchical processing from acoustic to semantic representations, where early layers excel at low-level tasks and alignment with primary auditory regions while late layers better reflect high-level tasks and alignment with late-language semantic regions. 

Recent work introduced the framework of brain-tuning (fine-tuning models using brain data as targets) for speech models \cite{moussa2024improvings}. It demonstrated that brain-tuning improves the semantic understanding of speech models, but it did not examine how brain-tuning affects the hierarchy of information processing in the model or how this hierarchy compares to the pretrained version. In this work, we utilize brain-tuning to investigate whether it could lead to models that reflect the speech processing hierarchy in the brain better than their pretrained counterparts. To this end, we build on the methods introduced in \cite{moussa2024improvings} to brain-tune popular speech models and then contrast the layer-wise brain alignment and downstream performance between the original pretrained models and the corresponding brain-tuned models. 

After testing this approach on two popular model families (Wav2Vec2.0 and HuBERT \cite{baevski2020wav2vec, hsu2021hubert}), we find that: 1) both brain alignment and downstream performance are generally improved in brain-tuned models (consistent with the findings in \cite{moussa2024improvings}). 2) Pretrained models have a downward trend (i.e., they peak in the middle) when aligning with high-level late language brain areas. This trend is similar to the one reported for the whole cortex and the auditory cortex in \cite{vaidya2022self}. Hence, upper-middle layers align with late language regions best while late layers perform worse, similarly to middle and early layers. 3) In contrast, brain-tuned models exhibit an upward trend, where late layers predict late language areas best. 4) Similarly to pretrained models, brain-tuned models have a downward trend when aligning with the low-level primary auditory regions. 5) For downstream tasks, phonemic and semantic tasks are best predicted by late layers in the brain-tuned models but by upper-middle layers in the pretrained models. As for low-level speech tasks, early layers perform best in both the pretrained and brain-tuned versions. 

Based on these findings, we propose that brain-tuned speech models serve as more suitable model organisms for human speech processing, offering both stronger performance and a hierarchy more consistent with human speech processing.

\section{Methods}
\subsection{Speech Language Models}
\label{method_used_models}
We analyze two popular pretrained self-supervised speech language model families: Wav2vec2.0 \cite{baevski2020wav2vec} and HuBERT \cite{hsu2021hubert}. Both models are transformer-based and rely on a CNN feature extractor to output latent speech representations that are fed to the transformer blocks. We use the base architectures, which were trained on $\sim$960 hours of audio in a self-supervised manner, where a portion of the input audio is masked, and the training objective is to predict the correct latent representations of the masked parts. Each model has 12 transformer layers and a fixed embedding dimension of 768. 

\subsection{Brain fMRI Dataset and Preprocessing}

\label{method_dataset}

We use the largest public fMRI dataset \cite{ds003020:2.2.0} in terms of the amount of data per participant, which provides 6.4 hours of fMRI recordings per participant. Each participant listened to 27 short stories from the Moth Radio Hour podcast while undergoing a whole-brain scan; there is a total of 8 participants in the dataset.  We use 25 stories of this dataset for brain-tuning and the remaining 2 stories as a held-out portion for
estimating brain alignment (refer to Sections \ref{method_btuning} and \ref{method_brain_encoding}
). We process this data by creating paired fMRI–audio snippets following \cite{oota2023speech, vaidya2022self, antonello2024scaling, moussa2024improvings}. This is done in three steps. First, a sliding window of $16\text{s}$ and a stride of $0.1\text{s}$ is applied to each audio recording. Second, we downsample the audio signal to match the slow fMRI acquisition rate using a three-lobed Lanczos filter (similar to \cite{ds003020:2.2.0, antonello2024scaling}). Third, we account for the delay in the hemodynamic response by modeling it as a finite response filter over 10 seconds \cite{ds003020:2.2.0, toneva2021bridging}. This process yields a paired audio snippets-fMRI dataset for brain-tuning and computing brain alignment.

\subsection{Brain-tuning}

\label{method_btuning}

For the speech model families mentioned in Section \ref{method_used_models}, we follow the brain-tuning process detailed in \cite{moussa2024improvings} to brain-tune each pretrained model. After constructing the dataset as in Section \ref{method_dataset}, given some input audio and its corresponding fMRI response, we fine-tune the pretrained speech model using the fMRI responses as targets, with the objective of reconstructing these fMRI responses (using an $L_2$ objective function). To achieve this, we add an average pooling layer on top of the output tokens followed by a projection head that maps the pooled output tokens to the corresponding fMRI response. We preprocess the fMRI responses by filtering out noisy voxels in the same manner as \cite{moussa2024improvings, antonello2024scaling}. The final set of voxels used for brain-tuning includes a substantial number of voxels from the late language regions and the auditory cortex. 

During training, we freeze the feature extractor and backpropagate the loss to fine-tune the projection head as well as the transformer layers. Since the number of voxels is variable among participants in the fMRI dataset, we brain-tune one model per participant; the final performance of brain-tuned models is then the mean and standard error across participants. 

\subsection{Brain Alignment}

\label{method_brain_encoding}

To evaluate the model’s ability to predict brain responses, we rely on the widely adopted brain alignment method used in \cite{antonello2024scaling, vaidya2022self, moussa2024improvings}. Brain alignment utilizes a linear function, which is trained to map a given layer’s representation of the input audio stimulus to the brain responses elicited by the same stimulus. The linear function is trained on a training portion of the dataset, using cross-validation, and is regularized using the ridge penalty. Brain alignment is evaluated by measuring how well the model and trained linear function can predict a held-out test set via Pearson correlation between the predicted and true test fMRI recordings\cite{moussa2024improvings, antonello2024scaling}. Then, we average this correlation for specific brain regions (e.g., the auditory cortex). This way, we obtain a scalar measure of how well different model layers can predict different brain regions. 

The final performance metric we report is the normalized brain alignment, where the brain alignment is divided by the data's noise ceiling (the maximum possible explainable variance) \cite{oota2023speech, antonello2024scaling, moussa2024improvings}. This provides a standardized measure for the alignment across different regions. Lastly, to focus our analysis on speech and language areas, we only report this normalized brain alignment for primary auditory and late language regions (e.g., angular gyrus, anterior and posterior temporal lobes, and middle frontal gyrus). These regions are parsed according to the labels and methods presented in \cite{oota2023speech}. We report this alignment measure individually for the different model layers, unlike \cite{moussa2024improvings}.

\subsection{Downstream Probing Tasks}

\label{method_downstream_probing}

Normalized brain alignment tells us about the ability of the SSL model features to predict brain data. To gain a finer-grained understanding of how linguistic representations evolve across these models, we test each layer's performance on known linguistic tasks. To this end, we linearly probe each layer’s representations to predict tasks with increasing semantic complexity: MFCC spectral features,  word identity, phonemes identity, and phonetic sentence type. These tasks span even more categories than the ones shown in \cite{vaidya2022self} and are carried out on datasets completely independent of the fMRI stimulus audio data, ensuring fair evaluation between brain-tuned and pretrained models. Below, we provide more information about each task.
\vspace{0.1cm}

\noindent \textbf{MFCC.} From the TIMIT dataset \cite{timit_asr}, we extract short audio frames and compute Mel-Frequency Cepstral Coefficients (MFCCs) by applying a mel-scale filter bank to the short-time Fourier transform of the signal, performing a logarithmic transformation, and then using the discrete cosine transform to obtain the MFCCs. We train a linear regression model on top of each layer’s representation to predict these MFCC coefficients. We measure performance by the coefficient of determination ($R^2$) on a held-out test set.
\vspace{0.1cm}

\noindent \textbf{Word Identity Prediction.} To assess the model’s ability to recognize words from audio, we train linear classifiers on top of each layer’s representation using the Speech Commands dataset \cite{commandds}. Each audio clip contains exactly one word chosen from a set of 35 possible commands. The classifier’s task is to identify which command was spoken, and we measure its performance by the F1-score on the test set.
\vspace{0.1cm}

\noindent \textbf{Phonemes Prediction.} We treat phoneme recognition as a multi-label classification problem, where a linear classifier maps the layer representation to a set of 39 possible phonemes that occurred in the original input audio segment. We use TIMIT \cite{timit_asr} for its phonetically rich sentences, and we report the classifier’s F1-score on the test set.
\vspace{0.1cm}

\noindent \textbf{Phonetic Sentence Type Prediction.} To evaluate a model’s phonetic understanding beyond single phonemes or words, we measure its performance in predicting the phonetic sentence type. In the TIMIT dataset \cite{timit_asr}, each sentence is labeled as one of three phonetic types: SA (designed to highlight dialectal variations and cover all English phonemes), SX (phonetically balanced with extensive phone coverage using fewer words), or SI (more natural and phonetically diverse). Each sentence type (SA, SX, SI) is designed to highlight specific phonetic, dialectal, or coverage aspects of speech. To predict the phonetic type of a sentence, the model must integrate phonetic patterns over an entire utterance, capturing both fine-grained acoustic details and broader contextual cues. We add a classification head to predict the sentence type from the layer’s representation. The final performance is the F1-score on the test set.



\section{Findings}

\begin{figure}[t]
  \centering
  \includegraphics[width=\linewidth]{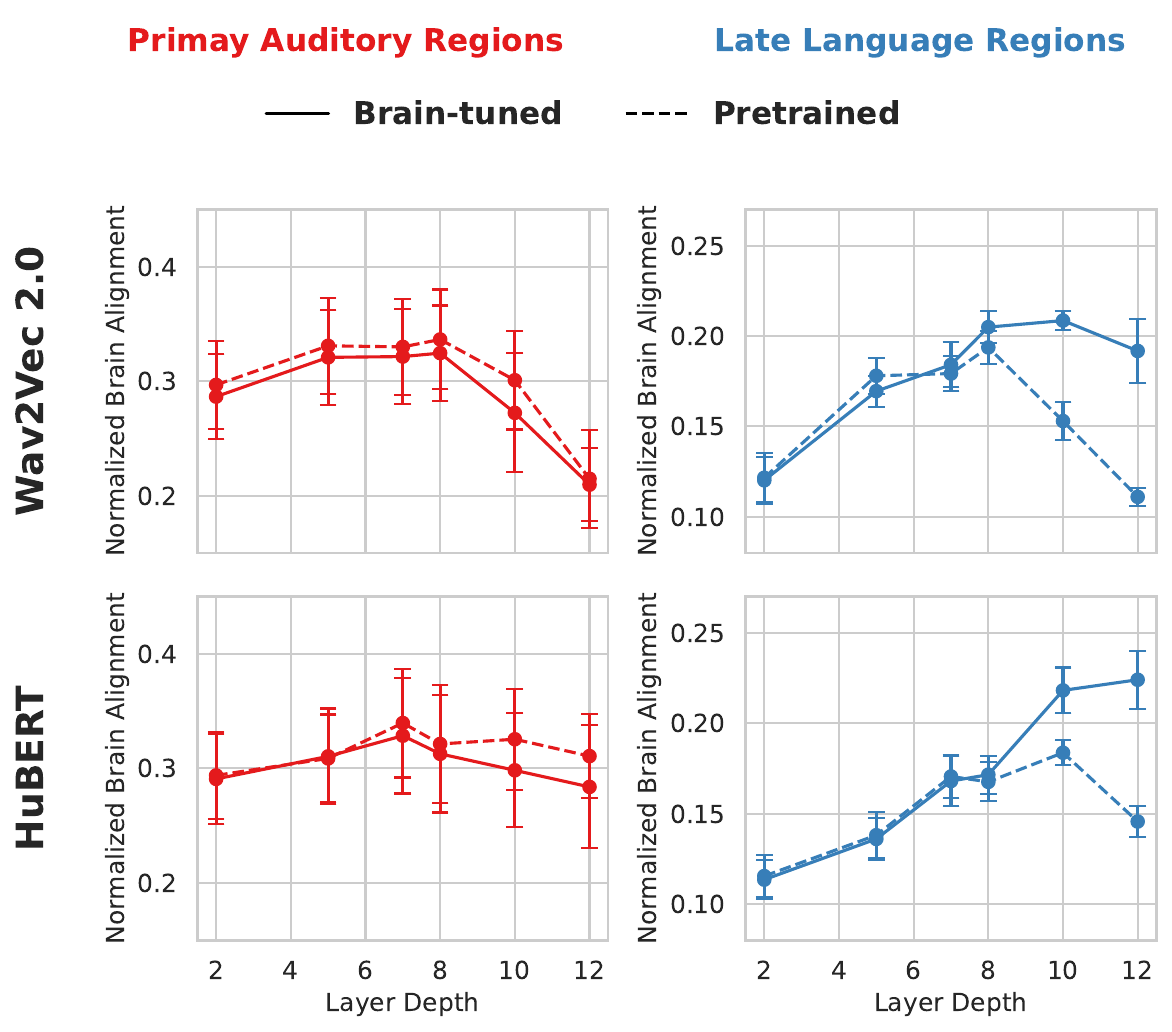}
  \caption{Layer-wise brain alignment for different regions. The alignment of late language areas increases substantially for the late layers of brain-tuned models, while no strong trend change is observed for the primary auditory areas. }
  \label{fig:res_brn}
\end{figure}

\begin{figure*}[t]
    \centering
  \includegraphics[width=\linewidth]{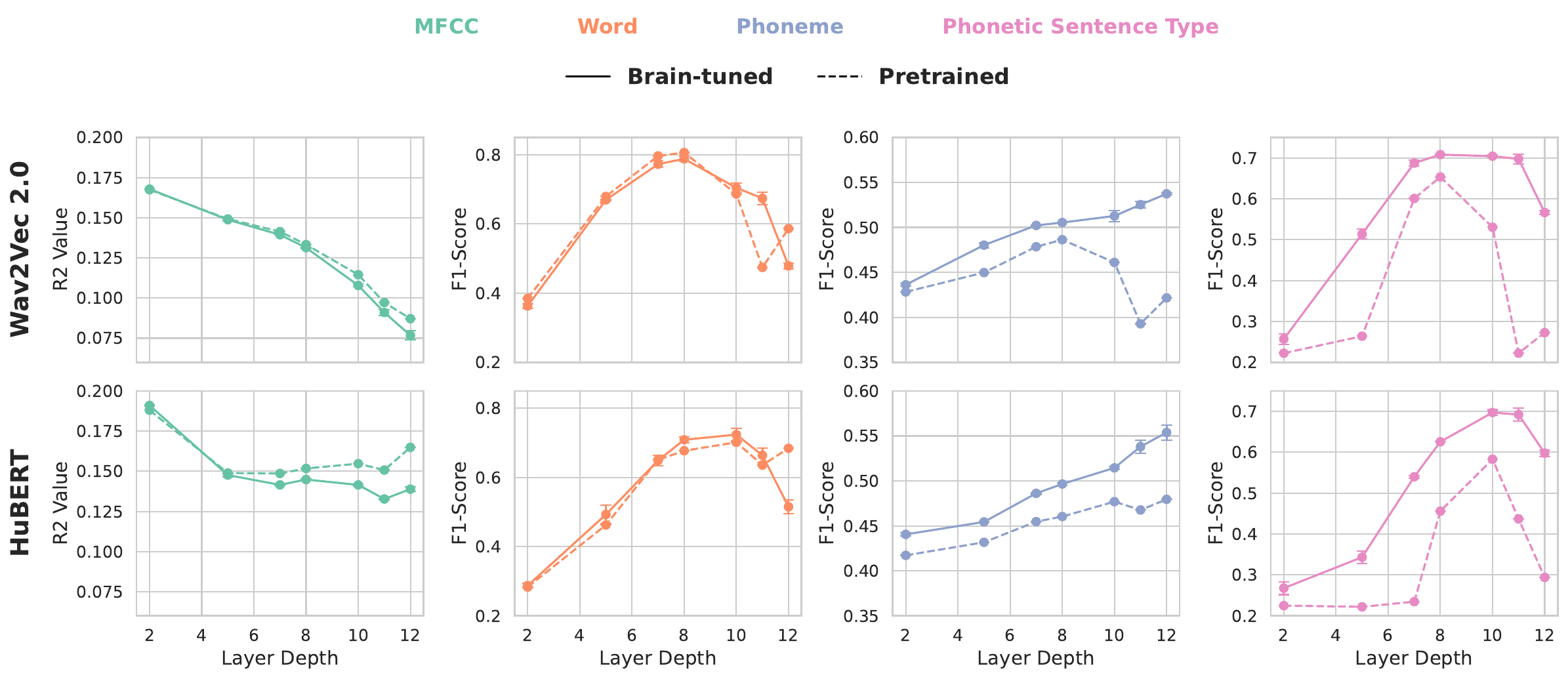}
  \caption{Layer-wise downstream performance. Late layers of brain-tuned models excel more in high-level tasks compared to their pretrained counterparts, while early layers excel in low-level tasks for pretrained and brain-tuned models. }
  \label{fig:res_downstr}
\end{figure*}

In this section, we present layer-wise results for brain alignment and linear probing for different downstream tasks with increasing semantics. These analyses investigate the hierarchy of information processing in brain-tuned vs. pretrained models.  

\subsection{Brain alignment results}
\label{results_brn}
We compute the normalized brain alignment described in Section \ref{method_brain_encoding} for two key language-related brain areas: the late language regions and the primary auditory regions. The late language regions are believed to support semantic language processing, while the primary auditory regions primarily handle lower-level processing related to the speech signal \cite{deniz2019representation}. For all models, we report the normalized brain alignment for both their pretrained and brain-tuned versions.


 In Fig.\ref{fig:res_brn}, we show the layer-wise mean and standard error of the normalized brain alignment across participants for all models. Overall, we observe that brain-tuning increases alignment with late language regions in both Wav2vec2.0 and HuBERT, and does not substantially impact the alignment with primary auditory regions. These observations are consistent with the brain alignment results of \cite{moussa2024improvings}, which were reported aggregately across participants and layers. More specifically, we observe the following:
\vspace{0.1cm}

 \noindent \textbf{Late language regions.} Brain-tuned models follow a mostly rising pattern across layers, where the later layers best predict late language regions. In contrast, pretrained models peak in the upper-middle layers, and their later layers perform similarly to their early layers. This trend for the pretrained models resembles the pattern observed for these same models over the whole cortex and the auditory cortex, as reported in \cite{vaidya2022self}.
\vspace{0.1cm}

\noindent \textbf{Primary auditory regions.} For the primary auditory regions, brain-tuned and pretrained models exhibit relatively similar performance, unlike for the late language regions. The highest alignment occurs around the middle layers, and the early layers are nearly on par with this peak. A notable difference is that brain-tuning appears to reduce alignment in the upper-middle and late layers, suggesting that it encourages these layers to capture fewer lower-level features.
\vspace{0.1cm}

Altogether, the differences in alignment with different regions between brain-tuned and pretrained models suggest that brain-tuning not only enhances the magnitude of alignment with semantic regions but also enforces a more brain-like hierarchy of information processing within the model. Specifically, later layers become substantially better at predicting responses in high-level language brain areas and notably worse at predicting responses in low-level speech brain areas compared to early and middle layers. Next, we further test the processing hierarchy via probing layers for tasks with different semantic levels.

\subsection{Downstream probing results}
To contrast how linguistic representations evolve across layers in brain-tuned versus pretrained models, we compare the models' layer-wise performance on the four tasks described in Section \ref{method_downstream_probing}. These tasks range from low-level features (MFCC) to more complex phoneme- and word-level features (Phoneme and Word Identity, Phonetic Sentence Type). We expect that a model with a more human-like hierarchy of speech processing across its layers would perform more complex tasks best with representations from late layers, and low-level tasks best with representations from early layers.

 In Fig.\ref{fig:res_downstr}, we present the layer-wise performance metric (detailed in Section \ref{method_downstream_probing}) for each task across all models. For brain-tuned models, we report the mean and standard error because brain-tuning is performed separately for each participant (resulting in \#participants brain-tuned models per model family). In general, brain-tuned models outperform the pretrained ones across all model families for more complex tasks, with notable gains in the Phonetic Sentence Type and Phonemes prediction tasks. When examining the layer-wise performance closely, we make the following observations on each task performance:
\vspace{0.1cm}

\noindent \textbf{Performance on low-level tasks.} MFCC spectral features are most accurately predicted by the early and middle layers in both brain-tuned and pretrained models. Notably, there is a slight reduction in performance at the late layers for the brain-tuned models, consistent with the decreased alignment in primary auditory brain areas observed in Fig.\ref{fig:res_brn}. This reduction, however, does not compromise the model's capability to do low-level speech tasks since the layers that encode these features best (the early layers) remain unaffected. Additionally, the magnitude of the drop is not considerable. 
\vspace{0.2cm}

\noindent \textbf{Performance on the word identity task.} Here, brain-tuned models perform similarly to pretrained ones, possibly because the pretrained models already perform very well. Moreover, we see no major changes in the hierarchy of brain-tuned models for this task. One possible explanation for this is that the pretrained hierarchy of word identity information is already ideal for the model, as it is beneficial for other tasks, such as phoneme prediction. This goes in line with suggestions from prior work that good phoneme perception benefits from word context rather than depending solely on acoustics \cite{vaidya2022self}. Thus, the best layers for word-level recognition may ideally precede the best layers for phoneme perception. Another possibility is that single-word recognition in this specific dataset may be possible using lower-level speech features, such as the number of phonemes, in which case it can be expected that late layers would not perform best at this task. Further testing on additional datasets can elucidate the brain-tuned models' behavior on this task. 
\vspace{0.1cm}


\noindent \textbf{Performance on phonetic tasks.} Beyond showing a marked improvement in predicting phonemes and phonetic sentence type, all brain-tuned models exhibit a clear upward trend for these tasks, with their late layers performing significantly better than the early and middle layers. In contrast, pretrained models follow a bell-shaped pattern, peaking at the upper-middle layers and dropping at the late layers —at times reaching similar performance to the early layers. This difference between pretrained and brain-tuned models tracks closely with the difference in brain alignment with late language regions in Fig.\ref{fig:res_brn}.
\vspace{0.1cm}

Overall, these results provide evidence that brain-tuned models exhibit a more brain-like hierarchical information processing than their pretrained counterparts: the early layers of brain-tuned models are the richest in low-level features, while the late layers are the richest in high-level features. In addition, brain-tuning not only enforces this hierarchy but also substantially boosts performance in more semantic tasks, in line with the results in \cite{moussa2024improvings}. The same behavior holds for brain alignment: brain-tuning improves the alignment of late layers with late language regions and maintains the alignment of early layers with early language regions (i.e., the auditory cortex).
\section{Discussion and Conclusion}
In this work, we utilize the method of fine-tuning speech models with fMRI data via brain-tuning \cite{moussa2024improvings} to show two pieces of evidence that brain-tuning reshapes speech models to more closely align with the hierarchy of human speech processing. 

First, brain-tuning improves alignment with late language regions for both Wav2vec2.0 and HuBERT, while leaving performance in primary auditory regions relatively unchanged (Fig.\ref{fig:res_brn}). This effect is most pronounced in the models' late layers, indicating that the models' later layers become more rich in high-level linguistic information. 

Second, brain-tuning leads to a notable shift in how linguistic representations evolve across layers. Specifically, early layers remain effective at capturing low-level spectral features, while later layers improve their representations for more complex phonetic and phonetic sentence-type tasks (Fig.\ref{fig:res_downstr}). This distribution of functionality aligns well with the brain alignment results (Fig.\ref{fig:res_brn}): primary auditory areas are best predicted by early model layers, whereas late language regions are predicted well by later layers. The downstream task evaluations confirm that brain-tuning enforces this hierarchical organization and enhances overall performance on more complex speech tasks. 


These results demonstrate that brain-tuned models can serve as better model organisms for speech processing in the brain. The previous best models (the pretrained speech models) lack a strongly defined semantic hierarchy: from Figs. \ref{fig:res_brn} and \ref{fig:res_downstr}, we see that the single best pretrained layer for late language regions is also the best for primary auditory regions. Moreover, the top-performing layers for high-level tasks perform on par with early layers for low-level tasks. This phenomenon—that the best layer covers spectrotemporal, articulatory, and semantic information—is also reported by previous work \cite{vaidya2022self}. Thus, we argue that models with a more well-defined semantic hierarchy can better serve as model organisms of listening in the brain as they reflect a clearer evolution of information and allow the model’s representations to encode richer and more specialized features, which is desirable, especially for small models. 

In conclusion, brain-tuning self-supervised speech models leads to better reflection of the speech processing hierarchy of humans. Aligning model representations more closely with brain data not only boosts performance on high-level downstream tasks but also establishes a consistently clearer progression from acoustics to semantics throughout the model layers. As a result, brain-tuned models emerge as more effective model organisms than their pretrained counterparts, which lag behind in both performance and hierarchical organization. Lastly, while the current brain-tuning method works very strongly on the tested model families, there is still room for improvement in boosting the performance and exploring how to enforce semantic hierarchy better in tasks like word identity recognition. Future work can also explore bigger models, broader datasets, and more tasks to further help reach performant models that align better with the auditory language processing in the brain.



\bibliographystyle{IEEEtran}
\bibliography{mybib}

\end{document}